\documentclass[conference]{IEEEtran}
\IEEEoverridecommandlockouts
\usepackage{cite}
\usepackage{amsmath,amssymb,amsfonts}
\usepackage{algpseudocode}
\usepackage{algorithm}
\usepackage{graphicx}
\usepackage{textcomp}
\usepackage{xcolor}
\usepackage{bm}
\usepackage{comment}
\usepackage{epsfig}
\usepackage{xspace}

\graphicspath{{./}{./figures/}}
\DeclareGraphicsExtensions{.pdf,.jpeg,.png,.jpg,.eps}

\DeclareMathOperator*{\argmax}{argmax}

\makeatletter
\def\onedot{\ifx\@let@token.\else.\null\fi\xspace}
\def\etal{\emph{et al}\onedot}
\def\Vec#1{{\boldsymbol{#1}}}
\def\Mat#1{{\boldsymbol{#1}}}
\def\wrt{w.r.t\onedot} 

\def\eg{\emph{e.g}\onedot}

\def\ie{\emph{i.e}\onedot} 

\def\BibTeX{{\rm B\kern-.05em{\sc i\kern-.025em b}\kern-.08em
T\kern-.1667em\lower.7ex\hbox{E}\kern-.125emX}}
\makeatother

\makeatletter
\def\footnoterule{\kern-3\p@
  \hrule \@width 2in \kern 2.7\p@} 
\makeatother

\begin{document}

\title{Learning Online for Unified Segmentation and Tracking Models\\
{}
}

\author{
  \IEEEauthorblockN{
    \hspace{1em}Tianyu Zhu\IEEEauthorrefmark{1} \hspace{2.1em}
    Rongkai Ma\IEEEauthorrefmark{1} \\
    Mehrtash Harandi \quad
    Tom Drummond%
  }
ARC Centre of Excellence for Robotic Vision, Monash University, Australia\\
\tt\small \{tianyu.zhu, rongkai.ma, mehrtash.harandi, tom.drummond\}@monash.edu

}

\maketitle
\begingroup\renewcommand\thefootnote{\IEEEauthorrefmark{1}}
\footnotetext{The authors contributed equally to this work.}
\endgroup

\begin{abstract}
Tracking requires building a discriminative model for the target in the inference stage. An effective way to achieve this is online learning, which can comfortably outperform models that are only trained offline. Recent research shows that visual tracking benefits significantly from the unification of visual tracking and segmentation due to its pixel-level discrimination. However, it imposes a great challenge to perform online learning for such a unified model. A segmentation model cannot easily learn from prior information given in the visual tracking scenario. In this paper,  we propose TrackMLP: a novel meta-learning method optimized to learn from only partial information to resolve the imposed challenge. Our model is capable of extensively exploiting limited prior information hence possesses much stronger target-background discriminability than other online learning methods. Empirically, we show that our model achieves state-of-the-art performance and tangible improvement over competing models. Our model achieves improved average overlaps of $\bm{66.0}\%$, $\bm{67.1}\%$, and $\bm{68.5}\%$ in VOT2019, VOT2018, and VOT2016 datasets, which are $\bm{6.4}\%$, $\bm{7.3}\%$, and $\bm{6.4}\%$ higher than our baseline. Code will be made publicly available. 
\end{abstract}
\section{Introduction}
\label{Introduction}

Visual tracking is an essential but complex task. It has a wide range of applications such as video surveillance \cite{tang2017multiple}, vehicle navigation \cite{lee2015road}, and human-robot interactions \cite{liu2012hand}.
A tracking algorithm must robustly discriminate the target from the background and provide an accurate estimation in the form of a bounding box or mask. 
Recent research shows that the pixel level discrimination offered by the unification of visual tracking and segmentation model boosts the estimation performance and improves the feature representation for classification ~\cite{siamMask, siamE}. 
Despite the success of such an approach, it suffers from two fundamental limitations. Firstly, little research has been done on adapting the segmentation model in online visual tracking applications. It is difficult for such models to fully exploit prior target information to improve target-background discriminability comparing to other online learning methods~\cite{henriques2014high,danelljan2018atom,nam2016learning}. More importantly, the second limitation is that the template image with the contextual background does not contain sufficient information about which object to mask. 
It takes an extensive hyperparameter search to achieve satisfying results to alleviate those two problems. 
\begin{figure}[t]
\begin{center}
   \includegraphics[width=1\linewidth]{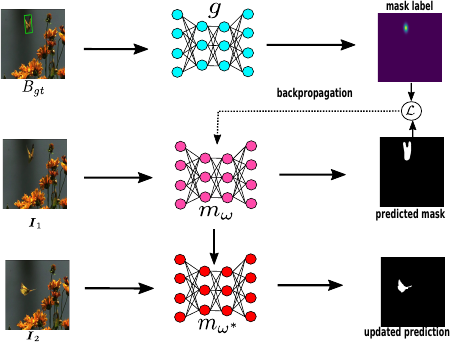}
\end{center}
   \caption{The adaptation process, as known as inner loop learning, of the segmentation head. A generator network $g$ is learned to generate the mask label $c$ from the bounding box $B_{gt}$ (coordinates). A loss based on the predicted mask and generated mask label backpropagates through the segmentation head $m_{\omega}$. As observed, the $m_{\omega^*}$ outputs a more accurate and informative segmentation.}
\vspace{-1.5em}
\label{fig:conceptual}
\end{figure}
In this paper, we propose a novel meta-learning approach where we train a model to learn from partial information. In the context of visual tracking, it can learn its segmentation model from the ground-truth bounding box given the first frame. With online adaptation enabled, it can break the limitations mentioned above. Moreover, the first frame of a video also contains rich information helpful for creating robust frameworks. 

We addressed two major challenges in this paper to enable online adaptation for unified segmentation and tracking model. Firstly, to perform one-shot learning for the segmentation model, a ground-truth must be given for the first frame. However, most tracking challenges only provide bounding boxes instead of a segmentation. We learn a generator to generate a mask label from the bounding box, which is used to aid the learning of the segmentation model, as shown in Fig.~\ref{fig:conceptual}. The generator is trained with the entire model using bi-level optimization. Secondly, the backbone of the state-of-the-art trackers usually benefits from a deep network (\eg, ResNet-50) that is computationally expensive to update online~\cite{siamMask}. In our approach, we propose to use a two-stage model where the initialization of network heads and body are optimized offline, and only heads are updated online. 
The main contribution of this paper is to propose a robust and unified  tracking and segmentation model that can perform adaptation through meta-learning from partial information (TrackMLP). Our ablation study shows that mere online adaptation hurts robustness while meta-learning without adapting the segmentation model does not improve the accuracy. Only with a constructive combination of both, our model outperforms the state-of-the-art algorithms in both accuracy and EAO for visual object tracking challenges~\cite{vot2018,vot2016,vot2019}.
\section{Related Work}

Early works often solely focus on the classification aspect of tracking using correlation filters~\cite{bolme2010visual,henriques2014high}, and deep learning feature extractors~\cite{siamfc,dcfnet} to produce a rough location of the object. In recent years, due to the demands from industry and benchmarks ~\cite{VOT_TPAMI,otb2013}, an increasing amount of research effort has been invested into the accuracy of bounding box estimation~\cite{danelljan2018atom,li2018high,wang2019spm}. Their approaches  usually benefit from the power of object detection models~\cite{maskrcnn,jiang2018acquisition,ssd} and fast siamese feature extractor~\cite{siamfc}. 
In this section, we briefly introduce the developments of visual tracking algorithms and meta-learning applications in tracking. 
\subsection{Visual Object Tracking}
Before deep learning methods gaining  popularity, correlation filters that can quickly adapt to the object during inference were the mainstream approaches for tracking models~\cite{bolme2010visual,henriques2014high}. Kernelized correlation filters~\cite{henriques2014high} particularly take advantage of non-linearity provided by the kernel trick while significantly boost the speed by exploit properties of circulant matrices. These traditional approaches are often simple, fast, and easy to implement. However, they cannot take advantage of large datasets available nowadays~\cite{muller2018trackingnet,ILSVRC15}. Moreover, the features captured by the kernels are limited in expressive-power compared to deep features.

SiamFC~\cite{siamfc} is the first work to utilize siamese networks and a correlation operation to produce a heat map indicating the location of the object. This algorithm is simple and relatively fast compared to other deep learning methods at the time. However, it has two significant problems. Firstly, the final bounding box has a fixed shape and only a few scales that do not consider the deformation of the object. To solve this problem, 
a region proposal network is used to take the result of correlation as input and generates more accurate bounding boxes~\cite{li2018high}. SiamMask~\cite{siamMask} takes one step further and learns to generate a mask of the object from video segmentation data Youtube-Vos~\cite{xu2018youtube}. During the inference stage, it is able to generate polygons from the masks, which significantly boosts the accuracy.
The second problem is due to its offline-only training strategy. As a result, it cannot distinguish similar objects and often classifies background as an object since the background can occupy a significant portion of the template image patch. SiamMask suffers from the same problems. More involved works such as~\cite{danelljan2018atom} using modulation vectors and gradient descent approximations to adapt the particular object during inference. This work, however, does not have online adaptations for the bounding box estimation. In addition to that, Bhat~\etal~\cite{bhat2019learning} have been proposed to utilize richer background information and previous frames to solve the latter problem.  

\subsection{Meta Learning for Tracking}
Tracking can be framed as a one-shot learning problem. The algorithm has one chance to see the ground-truth of the object in the first frame, and then it is required to estimate the location and shape of the object in the following frames. Meta-learning algorithms are shown to be useful for one-shot or few-shot learning problems~\cite{maml,nichol2018reptile}. Instead of updating the pre-trained weights during the inference stage, we can use the meta-trained initialization trained to adapt to a specific task. If the optimal sets of parameters for different tasks live in different manifolds, then the goal of meta-training is to find a set of parameters close to all these manifolds. 

MDNet~\cite{nam2016learning} is a visual tracking framework that uses a shared weight backbone network to generate generic features and a video-specific final layer as a discriminator to adapt to the object. The video-specific layer is randomly initialized during inference. The accuracy and speed of the process can be improved by learning an initial set of weights that can quickly adapt to the object~\cite{park2018meta}. Although performance improved, the model still needs to have significant online updates comparing to the siamese framework since the siamese framework has direct access to the template features. At the same time, MDNet must encode this information into the parameters of the network. Therefore, it makes sense to apply the meta-learning strategy to siamese frameworks. Choi \etal~\cite{choi2017deep} use a meta-learner model that takes gradients of the final layer as input and outputs new kernel weights and channel-wise attention to adapt to the new object. This is shown to be useful, but it still suffers from the inability to produce accurate bounding boxes inherited from SiamFC~\cite{siamfc}. We leverage the meta-learning, siamese framework, and expressive power of the segmentation model to achieve better performance in our work. 
\section{method}
\label{method}
\begin{figure*}[t]
\begin{center}
\scalebox{0.8}{
  \includegraphics[width=1\linewidth]{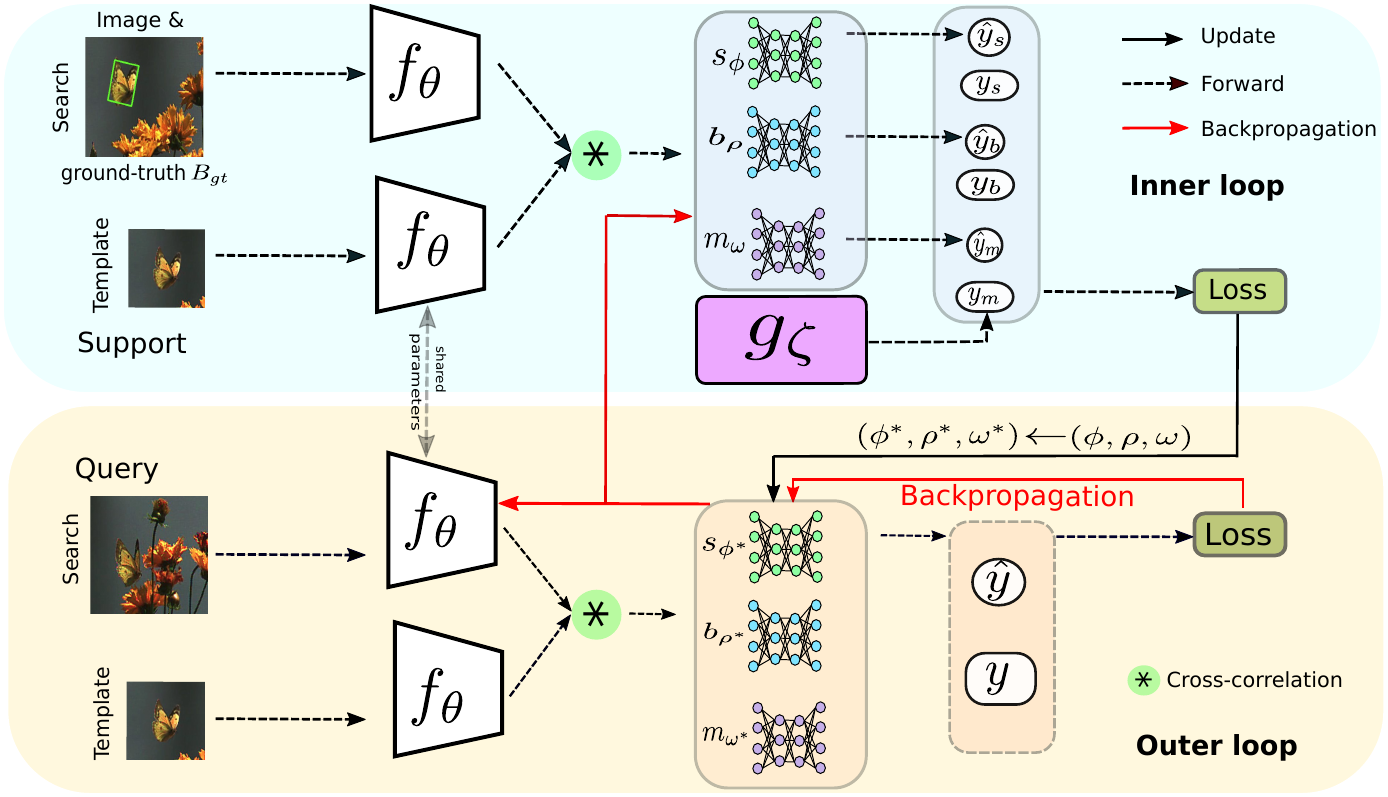}
}
\end{center}
    \caption{Our proposed TrackMLP framework. During the inner loop of learning, a set of updated weights $(\phi^*, \rho^*, \omega^*)$ is obtained using the supervision signal from support set pairs and generated mask labels. The computational graph is retained. For outer loops, Query pairs and real masks are used to compute loss, which is backpropagated through the adaptation process to optimize initial parameters $(\phi, \rho, \omega)$ and backbone feature extractor. The input of the heads is the resultant of the channel-wise correlation of two feature blocks extracted from template and search patches. All the backbones share the parameters.}
\label{fig:Hero}
\end{figure*}

In this paper, we present TrackMLP, a unified segmentation and tracking model that possesses online learning ability. This section first describes our two-stage meta-learning for model heads and backbone, followed by the introduction of meta-learning from partial information for the segmentation model in the tracking scenario.

\subsection{Meta-learning approach}
\subsubsection{Model Setup}
Our framework consists of a feature extractor followed by three functional heads. From here on, we denote the backbone network by function $f_{\theta}$ with parameters $\theta$, the classification head by function $s_{\phi}$ with parameters $\phi$, the localization head by function $b_{\rho}$ with parameters $\rho$, and the segmentation head as function $m_{\omega}$ with parameters $\omega$. Moreover, we use $h_{\varphi}$ and $\varphi$ to denote a head and its parameters in general (See Fig.~\ref{fig:Hero} for more details).
The input of the model consists of a template image with a smaller size (\eg, $127 \times 127$) and a search image (\eg, $255 \times 255$). The feature extractor first encodes both the template and search images into feature space, and then a channel-wise correlation is computed between the feature blocks (\ie, $r = f_{\theta}(z)\ast f_{\theta}(x)$). The resultant is then used as the input of the three functional heads.

\subsubsection{Construct Support and Query Set}
\label{subsubsection: Construct Support and Query Set}
Meta-learning process often requires the construction of support and query sets~\cite{maml,lifchitz2019dense,nichol2018reptile}. Formally, a task consists of a support set (\ie, $\mathcal{S} = \{ (\Mat{X}^s_{j}, y^s_{j})| j = 1, \ldots, K\}$), where $\Mat{X}^s_{j}$ denotes the $j$-th sample that contains the object of interest and $y^s_j$ denotes the corresponding bounding box, and a query set (\ie, $\mathcal{Q} = \{(\Mat{X}^q_{i}, y^q_i) |  i = 1, \ldots, N\}$), where $\Mat{X}^q_{i}$ denotes a query example. Both support and query examples contain the same object of interest. The model first learns from the support set, and the loss is computed based on the performance of the query set. The learning from the support set is practically an inner loop update, and  the update using the query set is the outer loop update. $\Mat{I}$ denotes a sequence of frames with length $L$, and a random frame $\bm{I}_{k}$ from the video is chosen to construct the support set. We crop both the template and search image patch from $\bm{I}_{k}$. Then $N$ additional search images are cropped from the rest of the images in the sequence, and the template image is cropped from $\bm{I}_{k}$ to construct the query set.  
The algorithm is expected to learn from the support set and its labels to improve its segmentation and tracking accuracy for the query set.

\subsubsection{Two-stage MAML}
\label{Two-stage MAML}

In the introduction,
we have argued the importance of online learning for tracking algorithms. Finn \etal~\cite{maml} have shown that a meta-learned initialization is superior to a set of pre-trained weights in terms of generalization ability, adaptation speed, and accuracy. Therefore, in this paper, we use the training process similar to MAML~\cite{maml} and Reptile~\cite{nichol2018reptile}. However, such an approach is less practical when a very deep network is used, and real-time constraints are considered in visual tracking. It takes a significant amount of time to do one iteration and needs more iterations to learn for deep networks. Therefore, we propose a two-stage MAML where only network heads are updated for the support set with one gradient step\footnote{Note that for simplicity, we use $\varphi$ to represent the parameters for a functional head in general. The adaptation for all the heads is the same.}, 
\begin{equation}
    {\varphi}^* = {\varphi} - \alpha \frac{1}{|\mathcal{S}|}\sum_{x \in \mathcal{S}}\nabla_{{\varphi}} \mathcal{L}_{\mathcal{S}} (h_{{\varphi}}(f_{{\theta}}(x))),
    \label{eq: Meta update}
\end{equation}
where $f$ is the backbone network with parameters $\theta$ and $h$ represents an arbitrary head of network with parameters ${\varphi}$. $\mathcal{S}$ and $\mathcal{Q}$ are the support set and query set, respectively. The step size $\alpha$ is a hyperparameter. Although only network heads are updated based on the support set, all the parameters are trained by optimizing the performance over the query set, including both ${\varphi}$ and ${\theta}$. The meta-objective is set as: 
\begin{equation}
    \min_{{\theta}, {\varphi}} \frac{1}{|\mathcal{Q}|}\sum_{x \in \mathcal{Q}} \mathcal{L}_{\mathcal{Q}} (h_{{\varphi}^*}(f_{{\theta}}(x))).
    \label{eq: Objective function}
\end{equation}
As shown by Eq.~\eqref{eq: Objective function}, the loss is computed using $\theta$ and $\varphi^*$, but the optimization is performed over $\theta$ and $\varphi$. Therefore the gradient of $\varphi^*$ is first computed, then the gradient of $\varphi$ is computed using the chain rule. 
With stochastic gradient descent, $\varphi$ is updated as follows:
\begin{equation}
    {\varphi} \leftarrow {\varphi} - \gamma \frac{\partial \mathcal{L}_{\mathcal{Q}}}{\partial {{\varphi}}^*}\frac{\partial{\varphi}^*}{\partial{\varphi}},
    \label{eq: Head grad}
\end{equation}
where $\gamma$ is the learning rate. As ${\varphi^*}$ depends on the derivative of $\mathcal{L}_{\mathcal{S}}$ over $\varphi$, the final gradient of ${\varphi}$ contains the second-order derivatives, which can be shown by: 
\begin{equation}
    \frac{\partial{\varphi}^*}{\partial{\varphi}}=1 - \alpha \frac{1}{|\mathcal{S}|}\sum_{x \in \mathcal{S}}\nabla_{{\varphi}}^2 \mathcal{L}_\mathcal{S} (h_{{\varphi}}(f_{{\theta}}(x)))).
    \label{eq: second order derivatives}
\end{equation}
Note that the gradients are computed across the support set and query set for ${\theta}$: 
\begin{equation}
    {\theta} \leftarrow {\theta} - \gamma \frac{\partial \mathcal{L}_\mathcal{Q}}{\partial h_{{\varphi^*}}}
    (\frac{\partial h_{{\varphi^*}}}{\partial f_{\theta}} \frac{\partial f_{\theta}}{\partial {\theta}} + \frac{\partial h_{{\varphi^*}}}{\partial {\varphi^*}} \frac{\partial {\varphi^*}}{\partial {\theta}}).
    \label{eq: Head grad}
\end{equation}
The gradient of ${\theta}$ also contains the second-order derivatives: 
\begin{equation}
    \frac{\partial {\varphi^*}}{\partial {\theta}} = -\frac{1}{|\mathcal{S}|}\sum_{x \in \mathcal{S}} \alpha \nabla_{{\varphi}} \nabla_{{\theta}} \mathcal{L}_\mathcal{S} (h_{{\varphi}}(f_{{\theta}}(x))),
\end{equation}
which is hard to compute for large networks. Therefore, we use a first-order approximation for the gradient of $\theta$.

\begin{figure}[h]
\begin{center}
   \includegraphics[width=0.90\linewidth]{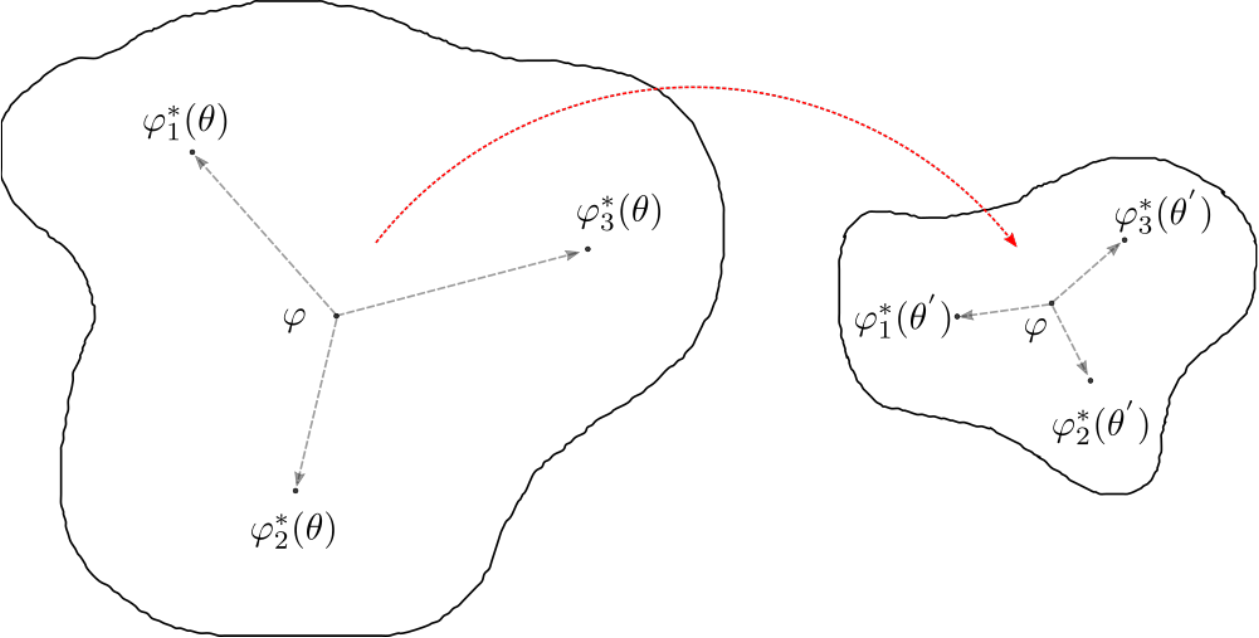}
\end{center}
\vspace{-1em}
   \caption{Illustration of our meta-learning process. Initial head parameter ${\varphi}$ is optimized to quickly adapt to new tasks with task optimal parameters $\varphi^*_1$, $\varphi^*_2$, $\varphi^*_3$. Feature extractor with parameter ${\theta^{'}}$ is optimized to make the head adaptation process even easier.}
\label{fig: Illustration of meta learning process}
\end{figure}

The intuitive effectiveness of the two-stage MAML is illustrated in Fig.~\ref{fig: Illustration of meta learning process}. The initial head parameters ${\varphi}$ is optimized to quickly adapt to new tasks. Task-specific head parameters (\eg, $\varphi_1^*$) is function of the backbone parameters ${\theta}$, as backbone maps input from image space to feature space, which is the input of the heads. Therefore, in two-stage maml, we aim to optimize ${\theta}$ and ${\varphi}$ in such a way that $\varphi$ adapts to new tasks even quicker. In the diagram, the distance between task optimal head parameters and initial head parameters are smaller with optimized backbone parameters (\ie, ${\theta}^{'}$). We find that a similar meta-learning approach demonstrated by another research~\cite{zintgraf2018fast} is also a great fit for tracking frameworks.
\vspace{-1em}

\subsubsection{Online Tracking}
\label{Online Tracking}
We augment the first frame to construct the support set  $\mathcal{S}$ during online tracking such that the samples in a batch can be highly correlated, which is a similar scenario as the training phase. Therefore, we set the BatchNorm layer of the functional heads (\ie, $s_{\phi}, b_{\rho}, m_{\omega}$) to evaluation mode (Please refer to~\S~\ref{Online tracking} for more implementation details). Moreover, the labels of classification and bounding box are calculated for each augmented search patch. More importantly, a mask is generated to provide training signals for the mask head. The generation of the mask label is discussed in the following~\S~\ref{subsection: Softmask}. We then compute the gradients by forward and backpropagating the support set and calculate new fast weights of the heads and store them in separate copies of heads. The rest of tracking procedure is performed using fast weights ${\phi^*}$, ${\rho^*}$, and ${\omega^*}$. Note that in previous mathematical derivations, we use ${\varphi}$ to present head parameters in general. Algorithm~\ref{alg: Online Adaptation} summarizes the process of online adaptation of our approach.

\subsection{meta-learning from partial information}
To track an object in a video, a ground-truth bounding box is provided for the first frame. Although this information is only a subset of the information provided by the pixel-level mask, it could still offer considerable constraint to improve the segmentation model. Such constraint cannot be exploited by the conventional meta-learning approach~\cite{maml,nichol2018reptile}. To this end, we propose a novel meta-learning method to enable learning from only partial information. 

In this approach, the inner loop update of the segmentation head is conditional on the bounding box coordinates $B_{gt}$. We use a generator network $g$, parameterized by ${\zeta}$, to generate the target label $c$ from source label $B_{gt}$. The generated $c$ is a $n^2$-dimensional joint distribution for a $n \times n$ image. During the inner loop learning, the adaptation loss for the segmentation model can be expressed as:

\begin{equation}
    c = g_{{\zeta}}(B_{gt}),
\end{equation}
\begin{equation}
    \mathcal{L}_{seg} = \sum_{ij} \log(1 + e^{-c_{ij}\hat{y}_{m_{ij}}}),
\end{equation}
where $c \in [0, 1]$ represents our generated mask and $\hat{y}_m$ is the output from segmentation head of the model. For this binary loss, the partial derivative \wrt~the output is
\begin{equation}
    \frac{\partial\mathcal{L}_{seg}}{\partial \hat{y}_{m_{ij}}}=c_{ij} \frac{-e^{-c_{ij}\hat{y}_{m_{ij}}}}{1+e^{-c_{ij}\hat{y}_{m_{ij}}}} .
\label{eq: grad seg}
\end{equation}
The outer loop training gradient of $g_{{\zeta}}$ can be computed as:
\begin{equation}
    {\zeta} \leftarrow {\zeta} - \eta\frac{\partial\mathcal{L}_{\mathcal{Q}}}{\partial{\omega^*}}\frac{\partial{\omega^*}}{{\partial c}}\frac{\partial c}{\partial\zeta},
    \label{eq: mask generator}
\end{equation}
where $\eta$ is a learning rate. Moreover, combining with Eq.~\eqref{eq: Meta update}, $\frac{\partial\omega^*}{\partial c}$ can be derived by chain rule as:

\begin{equation}
    \frac{\partial\omega^*}{\partial c} = -\frac{\alpha}{|\mathcal{S}|}\nabla_c\frac{\partial\mathcal{L}_{seg}}{\partial \hat{y}_m} \frac{\partial \hat{y}_m}{\partial\omega}
    \label{eq: generator grad}
\end{equation}
This can be further extended associated with Eq.~\eqref{eq: grad seg}.
The query set loss $\mathcal{L}_{\mathcal{Q}}$ is computed from the updated segmentation model and real ground-truth mask of the video segmentation dataset ~\cite{xu2018youtube}.           

\begin{figure}[h]
\begin{center}
   \includegraphics[width=0.85\linewidth]{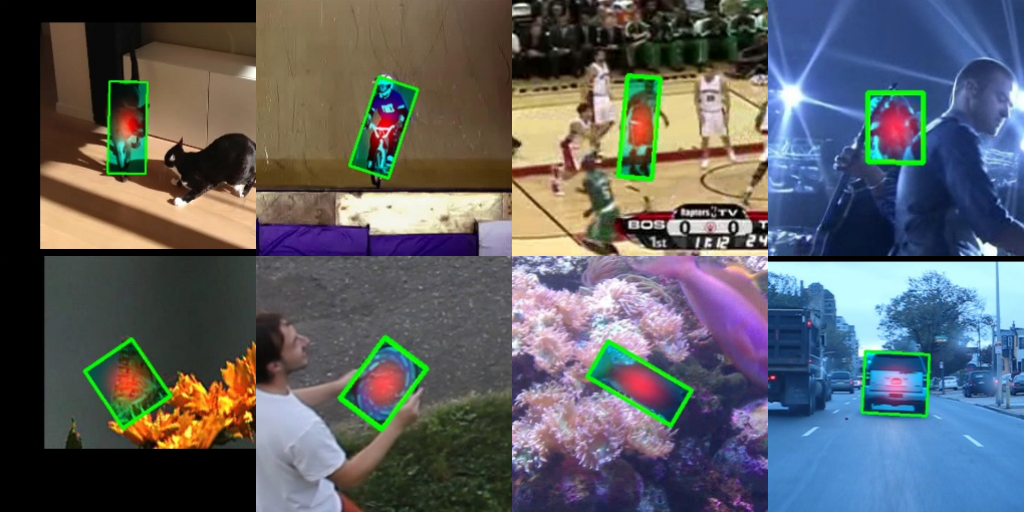}
\end{center}
   \caption{Generated masks label. To enable the online learning of mask head, we generate a soft mask label inside the ground-truth BBox using $g_\zeta$. All pixels outside of the BBox are labeled as negatives.}
\label{fig:SoftmaskExample}
\end{figure}

The generated mask labels are demonstrated in Fig.~\ref{fig:SoftmaskExample}. It acts as attention weights. The higher value of $c_{ij}$ means more likelihood that this pixel location is a positive example for segmentation. A lower value inside the bounding box but with a low-density value can be regarded as neutral, which does not incur much loss. Locations outside of the bounding box are labeled as negatives to incur high loss. 

\label{subsection: Softmask}

\begin{algorithm}[h]
   \caption{Online Mask Adaptation and tracking}
\textbf{Input:} An image sequence $\bm{I}$ with $L$ frames rotated Bounding Box $B_{gt}$ for initial frame

\textbf{Output:} Set of $L$ rotated bounding boxes $B$ and masks
\begin{algorithmic}[1]

\State Generate Support input $\mathcal{S}$ from $\bm{I}_1$, label $y$ from $B_{gt}$

\State ${\phi^*}$ $\leftarrow$ ${\phi} - \alpha\nabla_{\phi}\mathcal{L}(s_{\phi}(f_{{\theta}}(\mathcal{S}), y_s)$
\Comment{One step Adaptation}

\State ${\rho^*}$ $\leftarrow$ ${\rho} - \alpha\nabla_{\rho}\mathcal{L}(b_{\rho}(f_{\theta}(\mathcal{S}), y_b)$

\State ${\omega^*}$ $\leftarrow$ ${\omega} - \alpha\nabla_{\omega}\mathcal{L}(m_{\omega}(f_{\theta}(\mathcal{S}), y_m)$

\State Obtain a template image patch $z$ from $\bm{I}_{1}$ and $B_{gt}$
\For{$f=2$ to $L$}
\State Obtain a search image patch $x$ from $\bm{I}_{f}$ and $\bm{p}_{f-1}$
\Comment{$\bm{p}$ contains target position}

\State $\bm{p}_{f}\leftarrow \argmax_{{\phi^*}}{s_{{\phi^*}}(f_{\theta(z)},f_{\theta(x)})}$
\State Evaluate Masks $M = m_{{\omega^*}}(f_{\theta(z)}, f_{\theta(x)})$
\State Pick Mask at target position $M[\bm{p}_{f}]$
\State Generate  $B_{f}$ from $M[\bm{p}_{f}]$
\EndFor
\end{algorithmic}
\label{alg: Online Adaptation}
\end{algorithm}

\section{experiments}
\label{section: experiments}
To test the performance of our method, we evaluate the proposed method on visual object tracking benchmark VOT~\cite{vot2018,vot2016,vot2019}, generic object  tracking dataset GOT10K \cite{huang2019got}, and video segmentation dataset DAVIS~\cite{davis2016,davis2017}. To increase the reproducibility of our methodology, we also include training and testing implementation details. Furthermore, we conduct an ablation study to evaluate the effectiveness of each component of our model.  

\subsection{Training implementation details}
{\bf Datasets}.
\label{Datasets}
We use Youtube-vos~\cite{xu2018youtube} and Image-Net video detection~\cite{ILSVRC15} datasets for training. It is worthwhile to mention that only Youtube-vos~\cite{xu2018youtube} is used to train the segmentation head.  To leverage the meta-learning of our algorithm, we construct the support set and the query set. On top of the process mentioned in~\S~\ref{subsubsection: Construct Support and Query Set}, we augment the search image of the support set by flipping, blurring, and translation so that the support set contains a batch of $N$ augmented pairs (\ie, each pair consists of a search image and a template). Specifically, one batch of support set contains $4\times10$ examples from 4 different videos and 10 augmentations of one image from each video. The query set contains 10 different image samples from each corresponding video. The examples in one batch are highly correlated due to the nature of our training process, which violates the assumption of batch normalization~\cite{batchnorm,batchrenorm}. Therefore, we set the batch normalization layers to evaluation modes after pre-training. 

\noindent {\bf Network and update}.
We follow the good practice in~\cite{siamMask},
where ResNet-50~\cite{resnet} is used as our backbone
and three functional heads are all fully-convolution networks. 
We implement a 2-layer MLP 
(\ie, multi-layer perceptron) 
for the generator network (\ie, $g$). 
It is worthwhile to note that our generator network takes a two dimensional Gaussian prior produced by the ground-truth bounding box as input. Both the network backbone and heads are pre-trained on the datasets mentioned in~\S~\ref{Datasets} without the meta-learning process. 
We create identical copies for each head of the network. After the loss has been computed for the support set, we use the autograd function to compute the gradients with respect to the parameters of the heads. Then we use a fixed step size of 0.001 and store the updated parameters to the copies of the heads, and the graph is freed. When the query set is propagated, copies are used instead of original heads. 
After backpropagation is done, the gradients computed for copies are then loaded back to the original heads before optimizer stepping. Then we apply the gradients to finalize one iteration of the meta-update. For meta-update, we use SGD with a learning rate of 0.001 and momentum of 0.9. We use single Geforce GTX 1080 Ti for training. 
 
\subsection{Inference implementation details}
\label{Online tracking}
\noindent {\bf Augmentation}. The online support set is created using the first image, ground-truth bounding box, and generated soft mask. It is important to augment the support set. We randomly flip and shift the search patch to avoid location and orientation biases of heads and increase the number of examples. Scale variations and blurring are also applied. We use 40 augmented examples generated from the first frame as the support set. 

\noindent {\bf Adaptation}. For simplification, only one step inner loop update is done during one training iteration. During tracking, we experiment with a different number of steps and find 20 steps produce the best result. 

\noindent {\bf Mask model Variations}. 
Wang \etal
~\cite{siamMask} have experimented with both a small mask head and a refine-model, which is much bigger and contains skip connections. 
To have a fair comparison with the state-of-the-art result, we implement and outperform both model variations. The mask produced by the small mask head and refine-model have sizes of $63\times63$ and $127\times127$, respectively.
We upsample them to match the original scale. We have not tuned the hyperparameters such as segmentation threshold, window influence, and cosine penalty for online tracking.

\noindent {\bf Motion Model}. Our model outputs $5\times25\times25$ binary classification scores for 5 prior anchor boxes. It also outputs the same number of bounding box estimations and $25\times25$ mask estimations. We apply a cosine window to the classification scores to suppress similar objects, not close to the center. We find the location with the highest classification and use it to determine the center of the next search region, bounding box, and mask estimation. The size of the next search region is determined by the size of the bounding box estimation.

\noindent {\bf Hyperparameters}. We did not tune the hyperparameters including segmentation threshold, window influence and different penalties. We simply use the set of hyperparameters same as our baseline. However, it is possible to find out a set of better hyperparameters by exhaustive searching, which is out of the scope of this paper.

\subsection{Evaluation metrics}
\noindent {\bf Visual Object Tracking}. 
We evaluate our model on VOT datasets using accuracy, robustness, and expected average over laps (EAO). Accuracy measures average intersection over union between estimated bounding boxes and ground-truth bounding boxes while on track. Robustness equals the number of lost frames over total frame multiplying 100. EAO is a more complex metric where it measures overall performance, including both robustness and accuracy. For further information, we direct the readers to~\cite{vot2016}. Our approach is also evaluated on the GOT10K dataset\cite{huang2019got}. The SR, success rate shows the percentage of successfully tracked frames where overlap rates are above a threshold. For example,  SR$_{0.50}$ measures the number of frames where the overlap rates are above 0.5 over all the frames. 

\noindent {\bf Video Segmentation}. 
For DAVIS2017 and DAVIS2016, we simply measure mean intersection over union to show the effectiveness of our model. We generate a bounding box from the ground-truth mask of the first frame; the rest procedure is the same as VOT. We adapt our model to both the real mask and generated Gaussian soft mask to compare the performance. 

\subsection{Overall results}
\noindent {\bf Visual Object Tracking}.
In Table~\ref{tab: vot results}, we compare our method with the state-of-the-art models. In the first part of the result, our model has shown a clear advantage in accuracy without refine-model, hence overall EAO metrics in all three datasets. With the online adaptation to Gaussian soft mask, our mask head gains more information, achieving an accuracy gain of {\bf2.9}, {\bf3.5}, and {\bf2.9} percent gain from SiamMask. 

\noindent Refine-model with skip connections is a much bigger network head than the original mask head to perform more accurate segmentation but slower. SiamMask-E~\cite{siamE} takes the segmentation generated from the refine-model and fits an ellipse to generate the most accurate rotated bounding box we have seen during the period of this paper. To have a fair comparison with the state-of-the-art results, we implement our method with the refine-model and perform ellipse fitting. The result has shown that we can outperform the state-of-the-art models in accuracy and EAO on all three datasets. Moreover, Our model is more robust than them in VOT2018 and VOT2019. 

\begin{table*}[t!]
    \begin{center}
    
    \scalebox{1.2}{
    \begin{tabular}{l|c|c|c|c|c|c|c|c|c}
                            & \multicolumn{3}{|c|}{VOT2019} 
                            & \multicolumn{3}{|c|}{VOT2018} 
                            & \multicolumn{3}{|c}{VOT2016} \\
              
                            &  A$\uparrow$  & R$\downarrow$ & EAO$\uparrow$
                            &  A$\uparrow$  & R$\downarrow$ & EAO$\uparrow$
                            &  A$\uparrow$  & R$\downarrow$ & EAO$\uparrow$ \\
        \hline
        Atom*\cite{danelljan2018atom}    &  -            & -             & -
                            &  0.590        & 0.204         & 0.401
                            &  -            & -             & - \\
                            
        SiamFC++*\cite{siamfc++}           & -             & -             & -
                            &0.584          &$\bm{0.183}$          &0.426
                            &-              &-              &-\\
        
        SiamRPN++*\cite{li2019siamrpn++}           
                            &  0.595        & $\bm{0.467}$  & 0.287
                            &  0.601        & 0.234         & 0.415
                            &  0.642        & $\bm{0.196}$  & 0.464 \\

        SiamMask            &  0.552        & 0.502         & 0.260
                            &  0.555        & 0.267         & 0.348
                            &  0.587        & $\bm{0.196}$ & 0.420 \\
                            
        TrackMLP (ours)     &  0.625        & 0.512         & 0.295
                            &  0.633        & 0.258         & 0.410
                            &  0.659        & 0.210         & 0.455 \\
        
        \hline
        SiamMask-ref*       &  0.596        & $\bm{0.467}$  & 0.283
                            &  0.598        & 0.248         & 0.406
                            &  0.621        & 0.214         & 0.436 \\

        SiamMask-E-ref*     &  0.652        & 0.487         & 0.309
                            &  0.655        & 0.253         & 0.446
                            &  0.677        & 0.224         & 0.466 \\
                            
        TrackMLP-E-ref (ours)&  $\bm{0.660}$ & 0.497         & $\bm{0.310}$
                            &  $\bm{0.671}$ & 0.234         & $\bm{0.456}$
                            &  $\bm{0.685}$ & 0.228         & $\bm{0.478}$ \\
                         
    \end{tabular}
    }
    \end{center}

    \caption{Comparing with the state-of-the-art Siamese trackers on VOT2019, VOT2018, and VOT2016. Our tracker TrackMLP outperforms other trackers under the average overlap accuracy (A) and expected average overlap (EAO) metrics. $\uparrow$ stands for the higher the better, $\downarrow$ and  stands for the lower the better. * is for the numbers reported in the original paper. }
    \vspace{-1em}
    \label{tab: vot results}
\end{table*}

\vspace{2em}
\begin{table*}[h!]
    \begin{center}
    
    \scalebox{1.2}{
    \begin{tabular}{l|c|c|c|c|c|c}
                            
                            & {SiamMask}
                            & \multicolumn{4}{|c|}{}                    
                            & {TrackMLP} \\

        \hline

        Mask?
                            &\checkmark&\checkmark &\checkmark &\checkmark &\checkmark &\checkmark \\
        
        RPN Adaptation?
                            & &\checkmark &\checkmark & &\checkmark &\checkmark \\

        Rectangular Mask Adaptation?     
                            & & &\checkmark & & & \\

        Generated Mask Adaptation? 
                            & & &&\checkmark &\checkmark &\checkmark \\
                            
        Meta Trained Weights?   
                            & & & & & &\checkmark \\
        \hline
        VOT2018 Accuracy    
                            &0.555 & 0.567 &0.556 & 0.577& 0.588& $\bm{0.633}$\\
                            
        VOT2018 Robustness    
                            & 0.267& 0.295& 0.286&0.276&0.300 & $\bm{0.258}$\\
                    
        VOT2018 EAO    
                            & 0.348 & 0.336 & 0.338 &0.359& 0.349& $\bm{0.410}$\\            
             
    \end{tabular}
    }
    \end{center}

    \caption{Ablation study. The results show that mere steps of online adaptation for the classification, and regression heads (RPN) does not improve the result much and risk losing robustness due to overfitting to the first frame. However, substantial improvement in accuracy can be achieved by mask adaptation. The overall performance can be further improved by using meta-trained weights. }
    \vspace{-1.5em}
    \label{tab: ablation}
\end{table*}

\noindent {\bf Generic Object Tracking}.
We also conduct experiments on GOT10K~\cite{huang2019got} dataset to evaluate our model. As the results in Table~\ref{tab: got10k} suggested, our model achieves the best performance over the baseline models in terms of the average overlap (AO), 0.5 success rate ($\text{SR}_{0.5}$), and 0.75 success rate ($\text{SR}_{0.75}$). It is worthwhile to note that we do not train our model on GOT10K dataset, which indicates that the performance on this dataset can be potentially boosted by having an extra training stage on it.

\begin{table}[H]

    \begin{center}
    \begin{tabular}{l|c|c|c|c|c|c}

                            & {AO$\uparrow$ }
                            & {SR$_{0.50}\uparrow$}
                            & {SR$_{0.75}\uparrow$} 
                            & {Speed$_{fps}\uparrow$}                \\              
                             
        \hline

        MDNet\cite{nam2016learning}                           
                                       &0.299 &0.303 &0.099 &1.52 \\

        ECOOhc \cite{danelljan2017eco}        
                                       &0.286 &0.276 &0.096 &44.55 \\
                                
        SiamFC \cite{siamfc}
                                       &0.348 &0.353 &0.098 &44.15  \\
        
        SiamFCV2 \cite{valmadre2017end}
                                       &0.374 &0.404 &0.144 &25.81  \\

        SiamMask\cite{siamMask}   
                                       &0.464 &0.558 &0.206 &$\bm{48.71}$ \\ 
                                       
        TrackMLP (ours)
                                       &$\bm{0.474}$ &$\bm{0.570}$ &$\bm{0.226}$ &46.29 \\ 
    \end{tabular}
    \end{center}
    \caption{Comparing TrackMLP against baseline SiamMask \cite{siamMask} and other relevant trackers on GOT10K dataset\cite{huang2019got}. The results illustrate the efficacy of our algorithm TrackMLP based on the improvement of baselines. However, it is also applicable to other state-of-the-arts trackers due to the model-agnostic property.}
    \vspace{-2em}
    \label{tab: got10k}
\end{table}
\vspace{1em}
\noindent {\bf Video Segmentation}. 
Our model has reached mIOU of {\bf 0.565} for DAVIS2017 video segmentation dataset compared to 0.535 from SiamMask~\cite{siamMask}, which shows the effectiveness of our meta learned initialization.  
\begin{table}[t]
    \begin{center}
    \scalebox{1.0}{
    \begin{tabular}{l|c|c}
                            & DAVIS2017 
                            & DAVIS2016 \\
              
                            &  mIOU         & mIOU \\
        \hline
        SiamMask            &  0.535        & 0.712  \\
        
        TrackMLP-gm(ours)       &  0.560        & 0.721  \\

        TrackMLP-rm(ours)       &  $\bm{0.565}$ & $\bm{0.729}$\\
    \end{tabular}
    }
    \end{center}
    \caption{Davis Results. Our model outperforms SiamMask~\cite{siamMask} in mean intersection over union with either generated mask or real mask. This shows the effectiveness of our model.}
    \vspace{-2.5em}
    \label{tab: Davis results}
\end{table}

\noindent {\bf Speed}.
Once the model adapts to the first frame of video, the speed is the same as SiamMask~\cite{siamMask}, 85 fps. The adaptation process takes 0.9 seconds for TrackMLP and 1.5 seconds for TrackMLP-ref-E. It is possible to reduce adaptation time for the refine-model by meta-train an initialization that can involve fewer layers during online adaptation. There are also other methods focusing on reducing the adaptation time by using Gauss-Newton approximation~\cite{danelljan2018atom}. 

\subsection{Ablation study}
We perform an ablation study using a single dataset, VOT2018, to demonstrate the effectiveness of the method. 

\noindent {\bf RPN adpatation}. We find that direct online adaptation of classification and regression heads (RPN components) does not increase accuracy much and risks overfitting to the first frame hence reducing the robustness. This issue can be tackled by using meta-trained weights, as shown in Table~\ref{tab: ablation}.

\noindent {\bf Mask adpatation}. Substantial improvement in accuracy can be achieved by directly adapting the segmentation head to the first frame. We show that the generated mask as the supervision signal outperforms the rectangular mask with the same shape as the bounding box. 

\noindent {\bf Meta learned weights}. The experiments show that direct online adaptation improves accuracy but hurts the robustness. The meta learned initialization is essential to solving this problem. Moreover, the accuracy can be significantly further improved by using meta trained initialization. In conclusion, generated mask adaptation and meta trained weights as a combination significantly improve overall performance. A lack of either component results in minor improvement. 
\subsection{Qualitative analysis}
\begin{figure}[t!]
\begin{center}
\scalebox{1.0}{
   \includegraphics[width=0.95\linewidth]{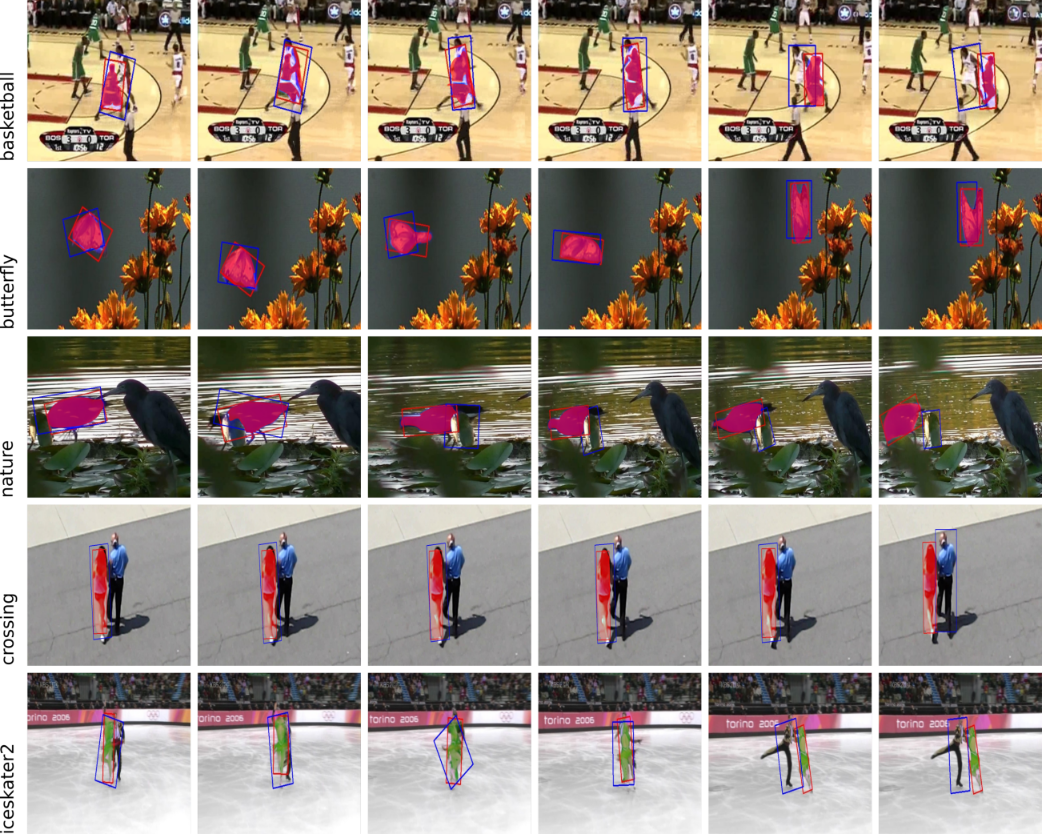}
}
\end{center}
   \caption{Qualitative examples. To illustrate the strength of our mdoel (TrackMLP) over the state-of-art tracking model, we pick up 30 sequential frames from 5 different vidoes of VOT2018 dataset, where the blue box is from SiamMask and the red box is from ours. The red segmentation is predicted by our mask model (\ie, $m_{\omega}$). It is worth noting that instead of losing track like SiamMask, 
   our model can consistently track the object of interest.}
  
\label{fig:Qualitative}
\vspace{-1.5em}
\end{figure}

To visualize the performance, we compare our baseline SiamMask~\cite{siamMask}, which is a state-of-the-art algorithm, with our method TrackMLP in Fig.~\ref{fig:Qualitative}.
It is shown that SiamMask~\cite{siamMask} is prone to mask the distractors
(\ie, 
objects similar to the object of interest).  
SiamMask mistakenly tracks a different player as the target in the basketball video, while our model can consistently track the object of interest even though the distractors appear. More importantly, TrackMLP generally produces more accurate segmentation.

\section{Conclusion}
\label{section: conclusion}
In this paper, we introduce TrackMLP, a simple method to unify the advantages of both pixel-level segmentation and online adaptation for tracking. We formulate a two-stage meta-learning process tailored for online adaptation of siamese frameworks. We show that with this approach, our method can quickly adapt to the object of interest and outperforms the previous state-of-the-art in both accuracy and EAO. The segmentation model only requires a ground-truth bounding box initialization to perform online adaptation. For future research direction, we will investigate various other online adaptation techniques suitable for visual tracking problems and further optimize the efficiency of the adaptation process, which can potentially lead to fast online tracking.

{\small
\bibliographystyle{IEEEtran}
\bibliography{IEEEexample.bib}
}

\end{document}